\documentclass{article}
\usepackage{ijcai19}

\usepackage{times}
\usepackage{soul}
\usepackage{url}
\usepackage[hidelinks]{hyperref}
\usepackage[utf8]{inputenc}
\usepackage[small]{caption}
\usepackage{graphicx}
\usepackage{amsmath}
\usepackage{booktabs}
\usepackage{algorithm}
\usepackage{algorithmic}
\usepackage{multicol, blindtext}
\usepackage{color}
\usepackage{cleveref}
\newcommand{\idest}{{\it i.e.}, }
\newcommand{\exempli}{{\it e.g.}, }

\title{Image Captioning with Compositional Neural Module Networks}

\author{
Junjiao Tian
\And
Jean Oh
\affiliations
Carnegie Mellon University
\emails
\{junjiaot, hyaejino\}@andrew.cmu.edu}

\begin{document}

\maketitle

\begin{abstract}
In image captioning where fluency is an important factor in evaluation, \exempli $n$-gram metrics, sequential models are commonly used; 
however, sequential models generally result in overgeneralized expressions that lack the details that may be present in an input image. Inspired by the idea of the compositional neural module networks in the visual question answering task, we introduce a hierarchical framework for image captioning that explores both compositionality and sequentiality of natural language. Our algorithm learns to compose a detail-rich sentence by selectively attending to different modules corresponding to unique aspects of each object detected in an input image 
to include specific descriptions
such as counts and color.
In a set of experiments on the MSCOCO dataset, the proposed model outperforms a state-of-the art model across multiple evaluation metrics, more importantly, presenting visually interpretable results. Furthermore, the breakdown of subcategories $f$-scores of the SPICE metric and human evaluation on Amazon Mechanical Turk  show that our compositional module networks effectively generate  accurate and detailed captions.  
\end{abstract}

\section{Introduction}

The task of image captioning lies at the intersection of computer vision and natural language processing. Given an image, the task is to generate a natural language sentence describing the information conveyed in the input image. Image captioning has received increasing attention over the years. The prevalent encoder-decoder frame work~\cite{vinyals2015show} serves as the backbone of many derived models. \cite{lu2017knowing} introduced and refined the attention mechanism that allows for better feature extraction and interpretability. \cite{anderson2017bottom} further used Faster-RCNN~\cite{ren2015faster} to replace the fixed-resolution attention mechanism. Researchers~\cite{you2016image}~\cite{yao2017boosting} also found that high-level concepts can provide a more concise representation for an image.\\
The majority of existing approaches follows the sequential model where words in a caption are produced in a sequential manner--\idest the choice of each word depends on both the preceding word and the image feature. Such models largely ignore the fact that natural language has an inherent hierarchical structure. For example, each object can be associated with various attributes. Even with better feature representations and attention mechanisms, the sequential structure of these models tends to lead to 
generic descriptions that lack specificity.
The models~\cite{dai2018neural}~\cite{wang2017skeleton} exploring compositionality have been shown to produce more accurate, specific, and out-of-distribution sentences and perform well on SPICE~\cite{anderson2016spice}, a semantic metric . 
Compositional models, however, do not compare well to the sequential models on the $n$-gram metrics such as BLEU~\cite{papineni2002bleu}. Because semantic evaluation metrics such as SPICE tend to ignore fluency and assume well-formed captions, the $n$-gram metrics are still important in judging the fluency of the generated captions.
\begin{figure}
\centering
\includegraphics[width=8cm]{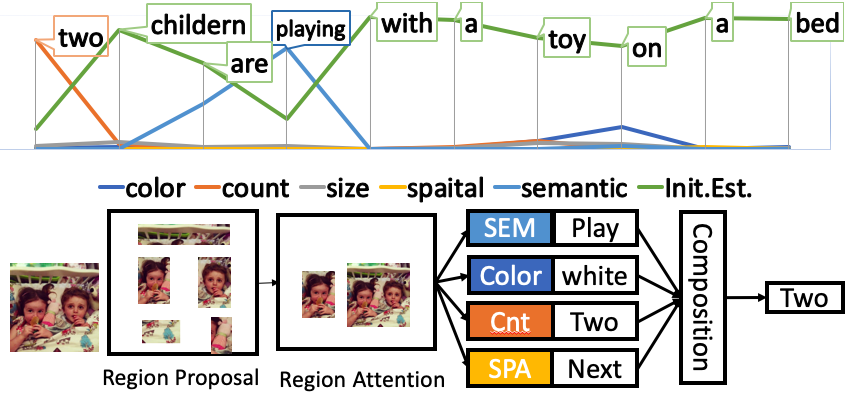}
\caption{Top: Visualization of attribute attention over time: the line plot shows one instance of time varying module attention. Note: Init.Est. stands for initial estimation. Bottom: 
An example of the workflow is shown in a diagram for time step 1 where the word ``two'' is generated. 
The model first chooses a region to focus on in the input image and the modules predict the attributes associated with the region. 
Note: SEM and SPA stand for semantic and spatial modules, respectively. }%
\label{fig:workflow}%
\vspace{0pt}%
\end{figure}

\vspace{0pt}
In this paper, we propose an image captioning model that combines the merit of sequential and compositional models by following a word-by-word generation process and combining \textit{grounded} attributes from \textit{specialized} modules. A high-level illustration of the workflow at one time step and visualization of the module attention is shown in~\Cref{fig:workflow}. More specifically, the algorithm first proposes regions of interest and then chooses a region to focus on depending on the context. The chosen region and the whole image are fed to a collection of functionally specialized modules where each module is delegated to predict one aspect of the objects such as count, color, and size. This is analogous to the Neural Module Networks (NMN)~\cite{andreas2016learning}, where each module is responsible for a specialized functionality and the final result is a dynamic composition of different modules. In our case, the model generates the final caption by dynamically attending to different modules.  The attributes, therefore, have a hierarchical dependency on and are grounded to the proposed regions.

With the proposed Compositional Neural Module Networks, we aim to generate detailed, specific captions without losing fluency, \exempli ``a red apple'' instead of ``a piece of fruit'' or  ``three people'' instead of ``a group of people.'' 
Overall, the main contributions of this paper are:
\begin{itemize}
    \item We 
    develop 
    a hierarchical model 
    that
    employs
    both \textit{compositionality} and \textit{sequentiality} of sentence generation. 
    \item Quantitatively, the model outperforms a state-of-the-art model on a set of conventional $n$-gram metrics and yields a noticeable improvement over the subcategories $f$-scores of the SPICE metric that is a more meaningful measurement of the semantics of generated captions. 
    \item Qualitatively, we perform human evaluation using Amazon Mechanical Turk. 
    According to the results, our model more often produces more detailed and accurate sentences when compared to the state-of-the-art model. A further analysis shows that the empirical results correlate positively with the quantitative results.
\end{itemize}

\section{Related Work}
In this section, we briefly introduce related and similar works and emphasize the differences of our model.

Most recent state-of-the-art models adopt the encoder-decoder paradigm, which is known as \textbf{NIC}~\cite{vinyals2015show}, where the image content is vectorized by a convolutional network and then decoded by a recurrent neural network into a caption. In this paradigm, attention-based models have been explored widely. For example, \textbf{AdaptAtt}~\cite{lu2017knowing} followed a top-down attention approach where an attention mechanism is applied to the output of the CNN layers. Another work~\cite{you2016image} used a word-based bottom-up attention mechanism.
\textbf{Top-Down}~\cite{anderson2017bottom} proposed a feature-based \textit{bottom-up attention mechanism} that retains spatial information whereas the word-based approach does not. Despite the improvements, these sequential models tend to generate words appearing more frequently and suffer from the lack of details in the generated captions.

~\cite{wang2017skeleton} presented a coarse-to-fine two-stage model. First, a skeleton sentence is generated by \textit{Skel-LSTM}, containing main objects and their relationships in the image. In the second stage, the skeleton is enriched by attributes predicted by an \textit{Attr-LSTM} for each skeletal word. \textbf{ComCap}~\cite{dai2018neural} proposed a compositional model, where a complete sentence is generated by recursively joining noun-phrases with connecting phrases. A \textit{Connecting Module} is used to select a connecting phrase given both left and right phrases and an \textit{Evaluation Module} is used to determine whether the phrase is a complete caption. In this work, noun-phrases are objects with associated attributes. In general, compositional models exhibit more variation and details in generated captions; however, they tend to perform poorly on the conventional $n$-gram metrics which are important measurements of fluency. 

Researchers have tried to explicitly model the compositionality of language in Question Answering (QA). This line of research shares a similar paradigm, namely, module networks. Module networks are an attempt to exploit the representational capacity of neural networks and the compositional linguistic structure of questions. ~\cite{andreas2016learning} learned a collection of \textit{neural modules} and a \textit{network layout predictor} to compose the modules into a complete network to answer a question. Rather than relying on a monolithic structure to answer all questions, the NMN can assemble a specialized network tailored to each question. We adopt this idea in QA to design a one-layer NMN with a collection of modules and a composition mechanism. Our model can compose a customized network depending on the context of a partially generated sentence.

\cite{you2016image} combined visual features with visual concepts in a recurrent neural network. \textbf{LSTM-A5}~\cite{yao2017boosting} also mined attributes as inputs to a language model. Although our model also uses attributes, the model differs fundamentally in several aspects. First, our model is hierarchical because attributes are grounded exclusively to selected regions that change over time. Second, the model is compositional because it combines grounded attributes and objects from separate detectors to predict the next word. Third, the attention is over the set of functionally specialized modules instead of individual visual concepts. Each module specializes in a single descriptive aspect of an object and determines the most probable attribute for that subcategory. For example, the color module generates different color predictions for different objects in an image depending on where the model's focus is. 

\section{Method}

\begin{figure}
\centering
\includegraphics[width=8cm]{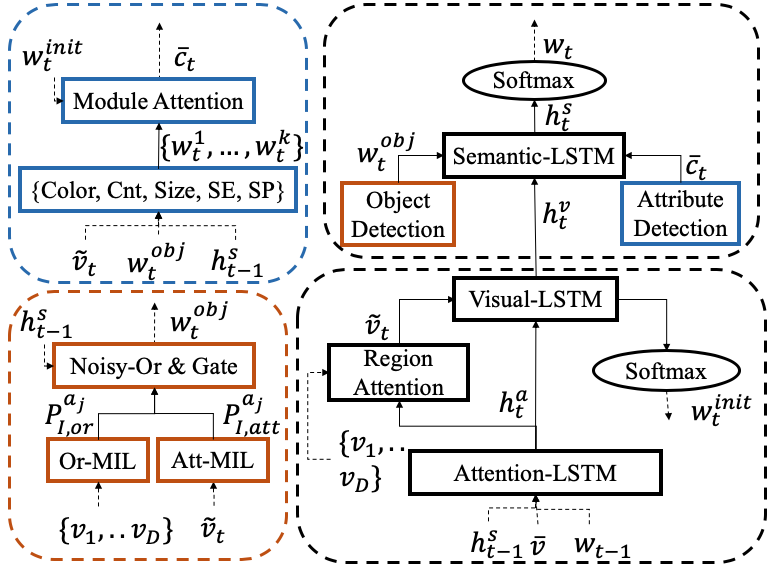}
\caption{Overview of the architecture: Right(Black): Recurrent Neural Network Trio, Top-Left(Blue) Modular Attribute Detection, Bottom-Left(Red) Stacked Noisy-Or Object Detection. Note: SE denotes Semantic and SP denotes Spatial.}
\label{fig:overview}
\end{figure}

The proposed hierarchical model for image captioning consists of three main components: Recurrent Neural Network (RNN) Trio, Stacked Noisy-Or Object Detection, and Modular Attribute Detection. We describe the overall captioning architecture as shown in~\Cref{fig:overview}, followed by technical details for the three components in~\Cref{sec:rnn-skeleton} --\ref{sec:modules} and the objective function used for training in~\Cref{sec:objectives}. 

Inspired by recent successes of the region-level attention mechanism ~\cite{anderson2017bottom}~\cite{yao2018exploring}, we use a Faster-RCNN in conjunction with a Resnet-101 backbone~\cite{he2016deep} to segment an image into a set of regions that likely contain objects of interest and encode each region $r$ as a fixed-length feature vector $\{v_1,..v_{D_r}\} \in R^{D_v}$ where $D_r$ is the number of regions, and $D_v$, the size of the feature vector. The feature vectors are used as inputs to other parts of the network.

The captioning model selects which region to attend to depending on the context. Given the region proposals, the stacked noisy-or object detection mechanism estimates all possible objects in the image regions. The modular attribute-detection mechanism operates on the attended regions to determine appropriate attributes for the attended object at each time step. The object and attribute detection makes up the compositional component while the RNN trio incorporates the detection results to generate a sentence in a sequential manner.

Similar to~\cite{anderson2016spice}, we divide the 
vocabulary into meaningful subcategories: 
an object set and five attribute sets which are color, size, count, spatial relationship, and semantic relationship. 
We select the six word-lists based on word occurrence frequency in the training data. The object set consists of visual nouns and the other attribute sets consist of adjectives. For example, \textit{red, green, blue} are in the color set and \textit{sitting, playing, flying} are in the semantic relationship set.

\subsection{Recurrent Neural Network Trio}\label{sec:rnn-skeleton}
The captioning model uses three recurrent neural networks, namely, \textit{Attention (A)-LSTM}, \textit{Visual (V)-LSTM} and \textit{Semantic (S)-LSTM}, to guide the process of generating captions sequentially. The input vector to the A-LSTM at each time step consists of the previous output of the S-LSTM, concatenated with the mean-pooled image feature $\bar{v} = \frac{1}{D}\sum_{i=1}^D v_i$ and encoding of the previous word. The attended image region feature, $\Tilde{v}_t$, is used as input to the V-LSTM to make an initial estimation of the next word based purely on visual evidence. In the final step, the information from the initial estimation, $h_t^v$, objects detection, $w_t^{obj}$, and attributes detection, $\bar{c}_t$, are combined to make the final prediction of the next word.

The attended image region feature $\Tilde{v}_t$ is obtained through the \textit{Region Attention} mechanism after the A-LSTM: 
\begin{align*}
    a_t &= \text{softmax}(W_b^T \text{tanh}(W_v V + (W_o h_{t-1})))\\
    \Tilde{v}_t &= \sum_{i=1}^D a_{t,i}v_i
\end{align*}
where $V \in R^{D_v \times D_r}$ is the set of image region features, $D_v$ is the dimension of visual features and $D_r$ is the number of region features. 
\subsection{Stacked Noisy-Or Object Detection}\label{sec:mil}
 Multi-label classification is a difficult task, where classes are not mutually exclusive in an image. 
 Here, we propose a stacked model that consists of two types of Multiple Instance Learning (MIL) object detectors to consider both image regions and the entire image simultaneously. 
 First, following the Noisy-Or Multiple Instance Learning (MIL) model used in~\cite{viola2016-mil-noisy}~\cite{karpathy2015deep}, we devise a 
 noisy-or detector to predict a distribution over a set of object labels. The noisy-or operation (\textit{Or-MIL}) is well suited to this task because it operates on each region separately and a positive detection from any region yields a high probability for the whole image. 
 Second, inspired by~\cite{ilse2018attention}, we adopt an attention based MIL (\textit{Att-MIL}) detector to consider the whole image, which contains large background objects such as ``grass.'' The two detection probabilities are combined with a second Noisy-Or operation, thus named the stacked approach. 
 
 Let us suppose that, for a given image $I$, there exist $V=\{v_1,v_2,...,v_{D_r}\}\in R^{D_v}$ image region features proposed by the Faster-RCNN network. The probability of an image containing  object $a_j$ is calculated by a Noisy-Or operation on all image regions of this image as follows:

\begin{align*}
    P_{I,or}^{a_j} = 1 - \prod_{v_i \in V} \left(1-p_i^{a_j}\right)
\end{align*}
where $p_i^{a_j}$ is the probability of object $a_j$ in image region $v_i$; $p_i^{a_j}$ is calculated through a sigmoid layer on top of the image region features. 


For the attention-based MIL detector, instead of an additional attention mechanism, we use the mean-pooled image region feature $\bar{v}$ as follows:
\begin{align*}
    P_{I,att}^{a_j} &= \frac{1}{1+e^{-f_{j,att}(\bar{v})}}
\end{align*}
where $f_{j,att}$ denotes parameters in a two-layer fully connected network. 

The final prediction, $P_{I}^{a_j}$, is computed using a second Noisy-or operation to combine the two probabilities $P_{I,att}^{a_j}$ and $P_{I,or}^{a_j}$:
\begin{align*}
    P_{I}^{a_j} &= 1 - \left(1-P_{I,or}^{a_j}\right)\left(1-P_{I,att}^{a_j}\right)
\end{align*}

We also design a gating mechanism to refine the object detection result at each time step. For example, if the word ``cat'' has already appeared in a sentence, we decrease its priority in the detection result for later time steps even though ``cat'' remains a positive instance for the image: 
\begin{align*}
    P_{I,t}^{a_j} = \text{relu} \left(W_h h_{t-1}^s + W_v \Tilde{v}_t\right) \circ P_{I}^{a_j}
\end{align*}
where { $ P_{I,t}^{a_j}\in R^{D_{obj}}$ } is the time-dependent prediction; $D_{obj}$, the size of the object set; $h_{t-1}^s$, the output of the S-LSTM at the previous time step; and $\Tilde{v}_t$, the attended image region feature at time $t$.

The output of the object detection module is a word-vector, $w_{t}^{obj}= E_{obj}P_{I,t}^{a_j}$,
where $ E_{obj}\in R^{D_{voc} \times D_{obj}}$ is a word embedding matrix from distribution over labels, $D_{obj}$, to the word-embedding space, $D_{voc}$. The word-vector $w_{t}^{obj}$ is used as an input to the S-LSTM for final decoding.

\subsection{Modular Attribute Detection}\label{sec:modules}
Attribute detection is achieved by using a collection of modules, each module $m\in M=\{m_1,...m_k\}$ with associated detection parameters $\theta_{m}$ and a \textit{Module Attention} mechanism to predict the layout of the modules. In this section, we describe the set of modules and the composition mechanism.

We use $k=5$ modules corresponding to different attributes of an object. They are: color, count, size, spatial relationship and semantic relationship modules. The modules map inputs to distributions over discrete sets of attributes. Each module has its own labels and, therefore, learns different behaviours.  

The modules all share the same simple architecture, a two-layer fully connected network. 
Customizing module architectures for different purposes might result in better performances as in~\cite{yao2018exploring} and~\cite{dai2017detecting}; in this paper, however,
we focus on the overall architecture and leave more sophisticated module architecture designs to future work. 
The distribution, $P_t^m$, over labels for module $m$ at time $t$ is computed using a softmax-activated function denoted by $f_m$: 
\begin{align*}
    P^m_t = f_m(\Tilde{v}_t,h_{t-1}^s,w_{t}^{obj}).
\end{align*}
The outputs of the modules are word vectors $w_t^m = E_m P_t^m$,
where $E_m$ is the word embedding matrix for module $m$. 

Next, we describe the compositional Module Attention mechanism 
that selects which module to use depending on the context. Inspired by \cite{lu2017knowing}, we use an adaptive attention mechanism and a softmax operation to get an attention distribution of the modules:
\begin{align*}
    z_t &= W_z^T \text{tanh}(W_m w^m_{t} + (W_g h_{t-1}^s))\\
    \alpha_t& = \text{softmax}(z_t)\\
    c_t &= \sum_{i=1}^k \alpha_{t,i}w_{t}^i
\end{align*}
where $w^m_{t} \in R^{D_{voc} \times k}$ is the module network outputs at time $t$. $k$ denotes the number of modules in consideration. We add a new element $w_t^{init} = E y_t^{init}$ to the attention formulation. This element is the word vector of the initial estimation of the next word from the V-LSTM.  
\begin{align*}
    \hat{\alpha} &= \text{softmax}\left([z_t;W_z^T \text{tanh}(W_iw_t^{init}  + (W_g h_{t-1}^s))]\right)\\
    \beta_t &= \hat{\alpha}[k+1]\\
    \Hat{c_t} &= \beta_t w_{t}^{init} + (1 - \beta_t)c_t
\end{align*}
Depending on the context, the network composes a different set of modules to obtain word-vector $ \Hat{c_t}\in R^{D_{voc}}$ for the S-LSTM.

\subsection{Objectives}\label{sec:objectives}
Our system is trained with two levels of losses, \textbf{sentence-level loss} and \textbf{word-level loss}. We first describe the more conventional sentence-level loss and then the auxiliary word-level losses.

\subsubsection{Sentence-Level Loss}
We apply two cross entropy losses to the V-LSTM and S-LSTM respectively:
\begin{align*}
    L_{V/S}= - \sum_{t=1}^T \log p(y_t|y_1,...,y_{t-1};I;\theta)
\end{align*}
where $\theta$ are the parameters of the models; $I$, the image; and $y=\{y_1,y_2,...,y_T\}$, the ground truth sequence. 

\subsubsection{Word-Level Loss}
We subdivide the word-level loss into two types: loss $L_{mil}^{att/or}$ to train the object and attribute detectors, and loss $L^{m}$ to train the module attention mechanism for composing attributes. 

\textit{Loss from Stacked Noisy-Or object detection:} as described in~\ref{sec:mil}, the MIL object detection has a stacked design. We train the noisy-or detector and attention-based detector using the two sigmoid cross entropy losses respectively:
\begin{align*}
    L_{mil}^{att/or} &= \sum_{a_j} - y^{a_j}\log(p^{a_j}) + (1-y^{a_j})\log(1-p^{a_j})
\end{align*}
where $y^{a_j}$ is $1$ when ground-truth object $a_j$ is present and 0 otherwise. $p^{a_j} \in \{P^{a_j}_{I,att},P^{a_j}_{I,or}\}$ is a sigmoid-activated function.

\textit{Loss from Modular Attribute detection:}
we use five masked cross entropy loss to train the attribute detection modules:
\begin{align*}
    L^{m} = \sum_{t=1}^T M_{t}^m \left(-y_t\log(P_{t}^m)+(1-y_t)\log(1-P_{t}^m)\right)
\end{align*}
where $m \in M$ and $M_{t}^m$ is $1$ if an attribute from set $m$ is present and $0$ otherwise at time $t$.

The composition mechanism is trained with the following additional loss:
\begin{align*}
    L_{c} = \sum_{t=1}^T M_t\left(y_{m,t}\log(\hat{\alpha})+ (1-y_{m,t})\log(1-\hat{\alpha})\right)
\end{align*}
where $M_t$ is $1$ if any ground-truth attribute is present and $0$ otherwise. $y_{m,t}\in R^{k+1}$ is a one-hot vector indicating which module is active at time $t$.

The final loss is a summation of all losses:
\begin{align*}
    L &=L_{V} + L_{S} + L_{mil}^{att}
    + L_{mil}^{or} + \sum_{m \in M} L_m + L_{c} 
\end{align*}
where $m\in M$ denotes an individual loss for each attribute module.

\section{Experiments}
\subsection{Datasets}
We use the MSCOCO dataset~\cite{lin2014microsoft} for evaluation. MSCOCO contains 82,783 training and 40,504 validation images; for each image, there are 5 human-annotated sentences.
We use the widely-used Karpathy Split~\cite{fang2015captions} to incorporate portion of the validation images into the training set. In total, we use 123,287 images for training and leave 5K for testing. As a standard practice, we convert all the words in the training set to lower cases and discard those words that occur fewer than 5 times and those do not intersect with the GloVe embedding. The result is a vocabulary of 9,947 unique words. 
For usage of the Visual Genome dataset~\cite{krishna2017visual}, we reserve 5K images for validation, 5K for testing and 98K images as training data. We refer the readers to~\cite{anderson2017bottom} for more details on training of the Faster-RCNN network.

\begin{table}[t!]
\centering
\begin{tabular}{p{1.9cm}rrrrr}  
\toprule
Model  & BL1 & BL4 & ROUGE & CIDER & SPICE \\
\midrule
NIC** &  - & 30.2 & 52.3 & 92.6 & 17.4 \\
AdaptATT** &  - & 31.2 & 53.0 & 97.0 & 18.1 \\
LSTM-A5** &  - & 31.2 & 53.0 & 96.6 & 18.0 \\
Top-Down**       & -   & 32.4 &53.8 & 101.1 & 18.7  \\
CompCap* &  - & 25.1 & 47.8 & 86.2 & 19.9 \\
\hline
Top-Down    & 76.7  & 32.0 & 59.0 & 105.4 & 19.9  \\
Ours:Complete  & \textbf{77.2 }& \textbf{33.0} &\textbf{ 59.4} &  \textbf{108.9} & \textbf{20.4 } \\
\bottomrule
\end{tabular}
\caption{Performance on the COCO Karpathy test split ~\protect\cite{fang2015captions}. Higher is better in all columns. * indicates results from the original paper. ** indicates re-implementation of the original papers by ~\protect\cite{dai2018neural}. Note: our implementation of the Top-Down model and the proposed model do not use beam-search whereas other results do. BL4$\/$1 denotes BLEU-4 and BLEU-1 respectively.}
\label{tab:metrics}
\end{table}

\begin{table}
\centering
\begin{tabular}{p{1.8cm}rrrrrr}  
\toprule
Model  & OBJ & ATTR & RE & CL & CT & SZ \\
\midrule
Top-Down    & 38.0  & 8.27 & 6.83 & 6.59 & 9.12 & 3.86      \\
Ours:Complete & \textbf{38.7} &\textbf{9.39} & \textbf{7.23} &\textbf{7.92} & \textbf{14.70} & \textbf{4.10} \\
\bottomrule
\end{tabular}
\caption{SPICE subcategory \textit{f}-score breakdown on the COCO Karpathy test split ~\protect\cite{fang2015captions}. Higher is better in all columns. Note the following abbreviations: OBJ-object, ATTR-attribute, RE-relations, CL-color, CT-count, SZ-size.}
\label{tab:breakdown}
\end{table}

\subsection{Implementation Details}
We set the number of  hidden state units in all LSTMs to 512, and the size of input word embedding to 300. We use a pre-trained GloVe embedding~\cite{pennington2014glove} and do not finetune the embedding during training. The pre-trained embedding is from a public website\footnote{\url{https://nlp.stanford.edu/projects/glove/}}  and consists of 6B tokens in total. In training, we set the initial learning rate as 5e-4 and anneal the learning rate to 
2.5e-4
at the end of training starting from the 20th epoch using a fixed batch size of 128. We use the Adam optimizer~\cite{kingma2014adam} with $\beta1$ to be 0.8. We train the Stacked Noisy-Or Object Detector jointly for 5 epoches and stop. The training is complete in 50K iterations.

To ensure fair comparison, we re-train the Top-Down using the same hyperparameters as the proposed model.
We report the results with greedy decoding to reduce the effect of hyperparameter search for different models.

We use the top 36 features in each image as inputs to the captioning models and do not finetune the image features during training.

\subsection{Amazon Mechanical Turk Setup}
\begin{figure*}
\centering
\includegraphics[width=17.5cm]{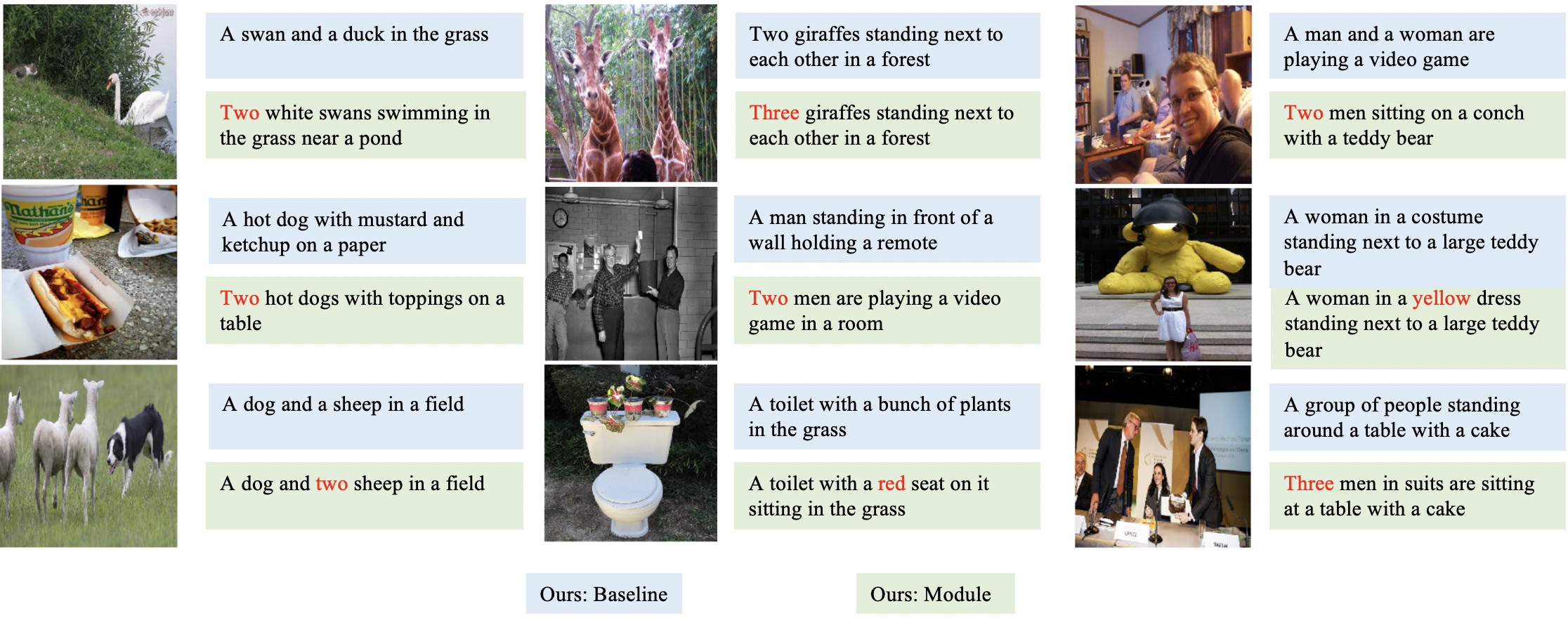}
\caption{Qualitative failed examples of captions generated by the Top-Down model (blue) and the proposed compositional module model (green). The proposed model makes mistakes in counting and associates color to the wrong objects in an image ocassionally.}
\label{fig:example_neg}

\includegraphics[width=17.5cm]{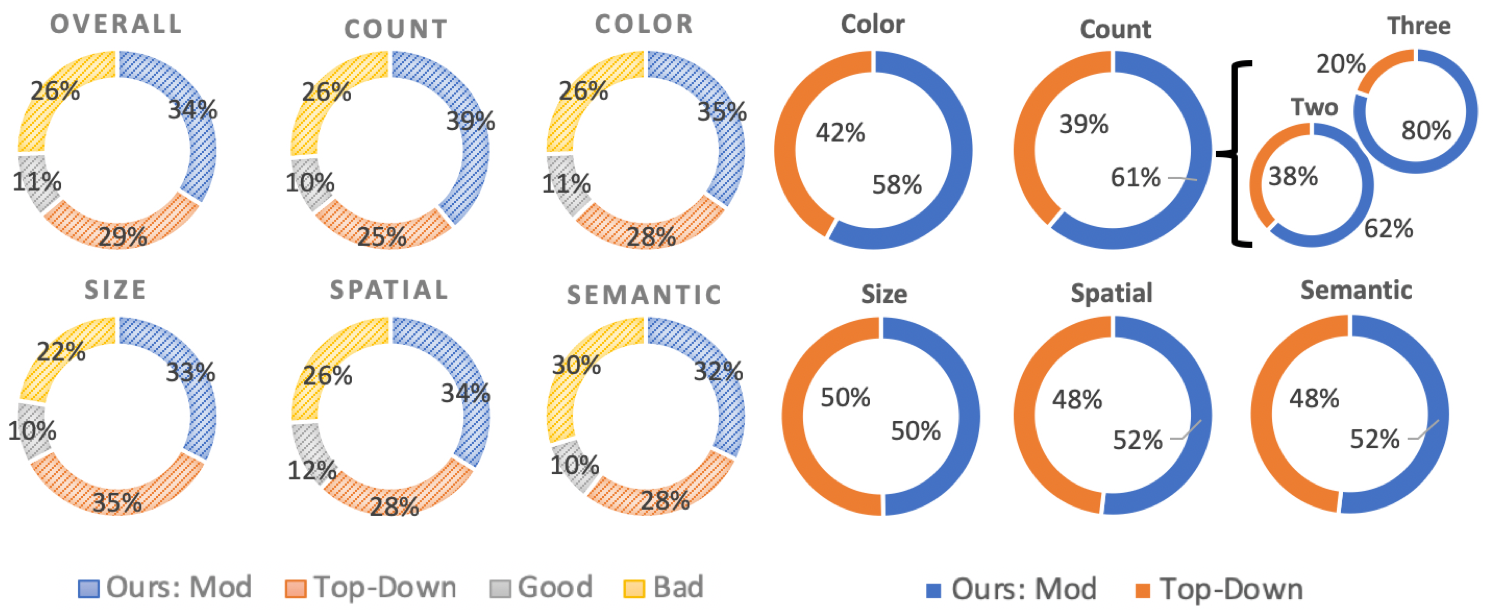}%
\caption{Left: Human evaluation results on the Caption Comparison task. The pie plot shows percentage of votes for different options. There are four options for participants, Option 1: \textit{caption 1}, Option 2: \textit{caption 2}, Option 3: \textit{equally good}, Option 4: \textit{equally bad}. Right: We count the number of occurrences of words from each subcategory word list in the 5K test split. The pie plot shows the ratio of word occurrences between the two models. We also show two specific examples from the count list, \exempli \textit{two} and \textit{three}.}%
\label{fig:AMT}%
\end{figure*}

Amazon Mechanical Turk (AMT) is a popular crowdsourcing service from Amazon. To investigate the effect of using compositional modules qualitatively, we design a Human Intelligence Task (HIT) to compare two captions generated from our implementation of the top-down model and the proposed compositional module networks. 
Each turker is asked to select from four options as shown in~\Cref{fig:AMT}: either of the two captions, equally good, or equally bad. For each image, we ask 5 workers to evaluate. 

For 1,250 images, 6,250 responses are received. The images are uniformly sampled from the test split; those images with identical captions from the two models are discarded. We design a qualification test to test workers' understanding of the problem and English proficiency. We adopt a max voting scheme to determine the quality of captions per image. 
When there is a clear winner,
we use it as the result for that image. 
In the case of ties, we give one vote to each tied option. 
\begin{figure*}
\centering
\includegraphics[width=17.5cm]{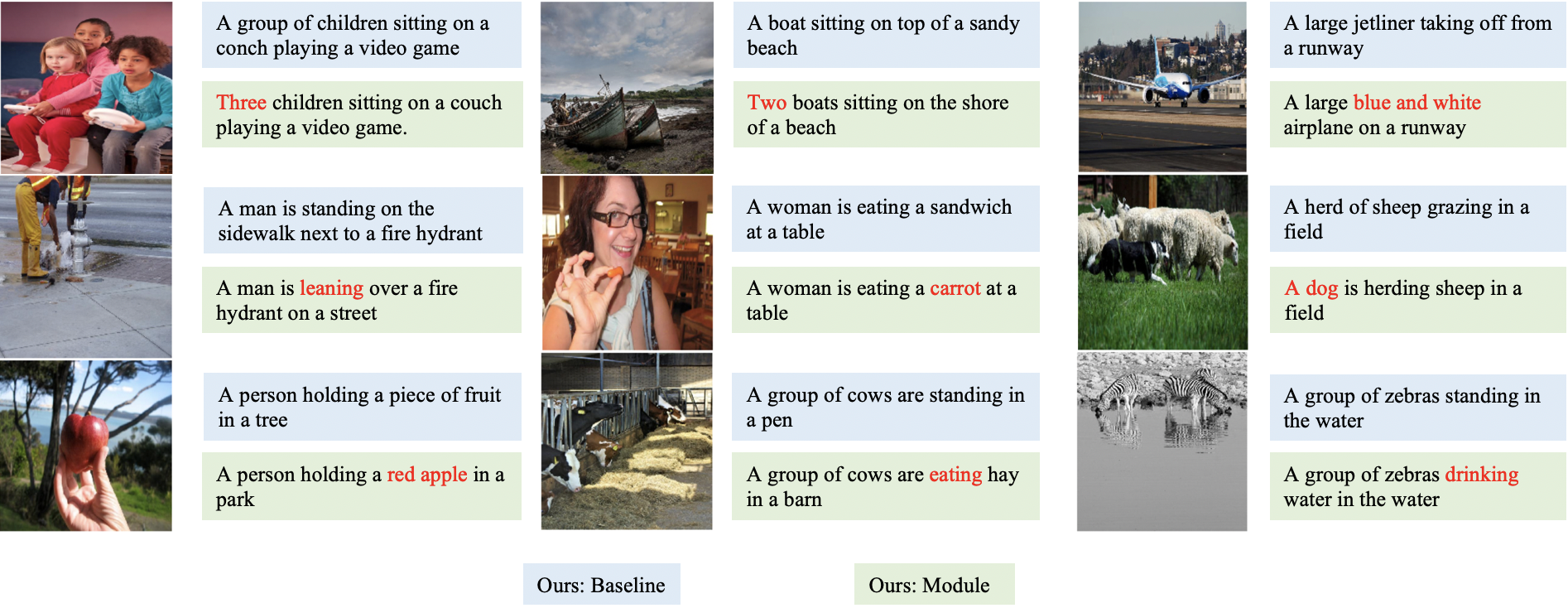}
\caption{Qualitative examples of captions generated by the Top-Down model (blue) and the proposed compositional module model (green). The proposed model produces more specific action attributes, \exempli ``leaning'' instead of ``standing,'' due to the semantic module.}
\label{fig:example}
\end{figure*}


\subsection{Experimental Results}
We compare our proposed model with our implementation of the \textbf{Top-Down} model~\cite{anderson2017bottom}, which achieved state-of-the-art performance on all evaluation metrics previously. We also list the published results of \textbf{CompCap}~\cite{dai2018neural}, which is 
another recent compositional model. We also include the published performance of \textbf{NIC}~\cite{vinyals2015show}, \textbf{AdaptATT}~\cite{lu2017knowing},  \textbf{Top-Down} and \textbf{LSTM-A5}~\cite{yao2017boosting} re-implemented by~\cite{dai2018neural} because the re-implementations use comparable visual features and are evaluated on the same test split. There are other models with better performances such as the model proposed by~\cite{yao2018exploring}, which uses additional datasets to train spatial and semantic relationship detectors. 
Our work is a fair comparison to the Top-Down model since both models use only MSCOCO as the main training data and Visual-Genome to train the Faster-RCNN, which is also used in~\cite{yao2018exploring}. 
%
Our implementation of the Top-Down achieves better performance than the implementation by~\cite{dai2018neural} and we use our implementation as the baseline for all comparison. 

Shown on the right side of~\Cref{fig:AMT}, a preliminary analysis of the generated captions shows that our proposed compositional module modle is able to generate captions that include more specific attribute words such as color and count. 
For example, the proposed model includes \textbf{4} times more 
of specific counts such as \textit{three} 
in its generated captions. 
 
\subsubsection{Evaluation Metrics}
We evaluate the approaches on the test portion of the Karpathy Split and compare the proposed approach against best-performing existing models using a set of standard metrics SPICE~\cite{anderson2016spice}, CIDEr~\cite{vedantam2015cider}, BLEU~\cite{papineni2002bleu}, ROUGE~\cite{lin2004rouge}, and METEOR~\cite{denkowski2014meteor} as in~\Cref{tab:metrics}. Our proposed model obtains significantly better performance across all $n$-gram based metrics. 

The $n$-gram metrics alone do not tell the whole story. We also report the performance on a recent metric, SPICE, and its subcategories \textit{f}-scores in~\Cref{tab:breakdown}. When compared to Top-Down, our module model achieves noticeable improvement on 
all subcategories but one.
The \textbf{count} subcategory is improved the most. We hypothesize that counting is an inherently difficult task for neural networks and sequential models tend to ``play safe'' by 
using generic descriptions instead.
This result demonstrates the effect of having dedicated functional modules for composition. It also shows that our proposed model can generate more detailed captions while improving fluency according to the $n$-gram metrics.

We also note that the \textbf{size} subcategory does not gain improvement over the Top-Down model. We hypothesize that this is due to the simple design of the module. Because the concept of size is a comparison between one object and its environment, our design only considers the object itself and the whole image. A more explicit representation of the concept of size such as bounding box might also be helpful.  

\begin{table*}[t!]
\centering
\begin{tabular}{lrrrrrr}  
\toprule
Model  & BL1 & BL4 & ROUGE & METEOR &CIDER & SPICE \\
\midrule
1. Up-Down &  76.7 & 32.0 & 59.0 & - &105.4 & 19.9 \\
2. Ours:w/o Mod    &   77.0 & 33.0 & 59.5 & 27.4 & 108.4 & 20.4 \\
3. Ours:w/o MIL  & 76.4   & 32.1 &59.0 & 27.2 & 106.1 & 20.1  \\
4. Ours:w/o (Mod+MIL)&  76.5 & 32.2 & 59.0 & 27.2 & 106.1 & 20.1 \\
5. Ours:w/o (Mod+AMIL)&   77.0 & 32.4 & 59.0 & 27.4 & 107.1 & 20.2 \\
6. Ours:Complete    & \textbf{77.2}  & \textbf{33.0} & \textbf{59.4} & \textbf{27.6} & \textbf{108.9} & \textbf{20.4}  \\
\bottomrule
\end{tabular}
\caption{Ablation study: Performance on the COCO Karpathy test split~\protect\cite{fang2015captions}. The higher, the better in all columns.}
\label{tab:ablation_ngram}
\end{table*}

\subsubsection{Ablation Study}
\begin{table*}
\centering
\begin{tabular}{lrrrrrrr}  
\toprule
Model &SPICE & OBJ & ATTR & RE & CL & CT & SZ \\
\midrule
1. Top-Down   &      19.9 & 38.0  & 8.27 & 6.83 & 6.59 & 9.12 & 3.86\\
2. Ours:w/o Mod    &   20.5 & 38.8 & 8.80 & 7.02 & 6.29 & 11.9 & 4.33 \\
3. Ours:w/o MIL  &     20.0 & 38.0 & 8.94 & 6.87 & 7.80 & 12.2 & 4.11 \\
4. Ours:w/o (Mod+MIL)& 20.1 & 38.3 & 8.20 & 6.89 & 6.03 & 9.23 & 4.32 \\
5. Ours:w/o (Mod+AMIL)& 20.2 & 38.5 & 8.64 & 7.01 & 6.51 & 9.37 & 4.45\\
6. Ours:Complete & \textbf{20.4}& \textbf{38.7} &\textbf{9.39} & \textbf{7.23} &\textbf{7.92} & \textbf{14.70} & 4.10 \\
\bottomrule
\end{tabular}
\caption{Ablation study: SPICE subcategory \textit{f}-score breakdown on the COCO Karpathy test split ~\protect\cite{fang2015captions}. Higher is better in all columns. Note the following abbreviations: OBJ-object, ATTR-attribute, RE-relations, CL-color, CT-count, SZ-size.}
\label{tab:ablation_spice}
\end{table*}

To show the effectiveness of each component, we conduct an ablation study on different variants of our model and compare the performance on SPICE $f$-scores and $n$-gram metrics. 
We use the following notations:
\textbf{Mod} stands for the modular attribute detectors; \textbf{MIL} stands for the stacked Noisy-Or object detectors; \textbf{AMIL} stands for the attention based MIL detector. For example, \textbf{Ours:w/o (Mod+AMIL)} is a model without modular attribute detectors and stacked MIL detector (but it has a single layer Noisy-Or detector).  

In~\Cref{tab:ablation_ngram}, comparing row 2 and row 6 shows that the modular attribute detectors do not contribute to the improvement on the $n$-gram metrics. Comparing row 4, 5, and 6 indicates that the MIL object detectors are the prime contributors for the improvements on those metrics (CIDEr 106.1$\rightarrow$107.1$\rightarrow$108.9) and our stacked design further improves the single layer Noisy-Or detector. 

In~\Cref{tab:ablation_spice}, comparing row 3 and 6, we can see that the MIL object detectors contribute to the object subcategory the most and also affects the performance on other subcategories a little. However, the absence of modular attribute detectors further deteriorated the performance on other subcategories, such as count (11.9$\rightarrow$14.0) and color (6.29$\rightarrow$7.95) when comparing row 2 and 6. 

In summary, the MIL object detectors contribute to the improvement on the $n$-gram metrics and object subcategory, while the attributes modules improve on other subcategories. The attribute detectors are responsible for improved semantics and object detectors are primarily responsible for improved fluency.

\subsubsection{Human Evaluation using Amazon Mechanical Turk}
We report the human judgment on the captions generated by the module model and the Top-Down model. As shown in~\Cref{fig:AMT}, $\textbf{5}\%$ more people prefer our model over the Top-Down. The difference becomes more significant when we consider subsets of the images. We split the evaluation set into subsets depending on whether their 5 ground truth sentences contain related attributes. For example, images in the Color subset contain at least one ground-truth sentence with a color attribute. The difference  is $\textbf{7}\%$ in the color subset and $\textbf{14}\%$ in the count subset. The result correlates with the largest improvement on the color and count subcategories in the SPICE subcategory f-scores. This highlights the 
strength
of our model in the subcategories. The human evaluation results 
qualitatively indicates that there is a discernible improvement recognized by human users. 

\begin{figure}
\centering
\includegraphics[width=8.5cm]{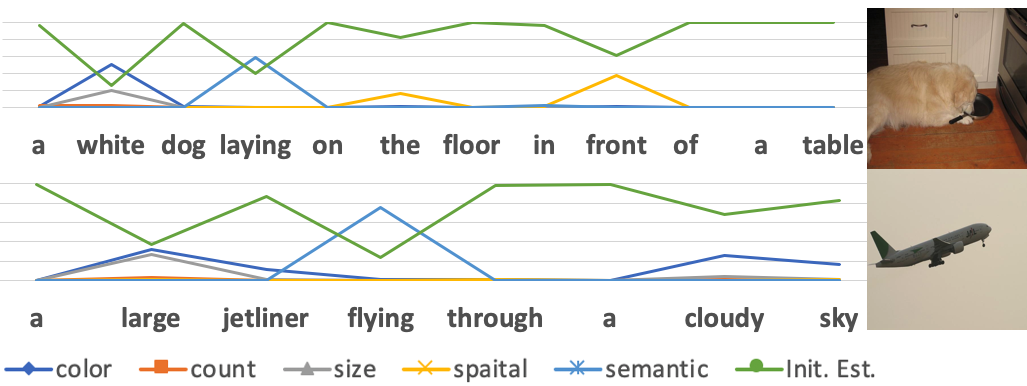}%
\caption{Interpretable visualization of Module attention over time. Note: Init.Est. stands for the Initial Estimation from the V-LSTM}%
\label{fig:attention}%
\end{figure}
\subsubsection{Qualitative Analysis}
\Cref{fig:example} shows sample captions generated by 
the Top-Down model and our proposed model.
The examples show that our model gives more accurate description of counting, color, and actions, \exempli more precisely describing a person's bent-over pose in the picture by using ``leaning'' instead of ``standing.''

\Cref{fig:example_neg} shows failure cases of the proposed model. Two most common mistakes are incorrect counting and color association. Occasionally, the proposed model tries to give a more specific description of counts of people/objects but the count is wrong whereas the baseline model uses a safe description such as ``a group of''; sometimes color is associated with a wrong object, \exempli our model predicts ``a woman in \textbf{yellow} dress'' whereas, in fact, the yellow attribute should have been associated with the teddy bear in the background. 
 
\Cref{fig:attention} shows two examples of changing module attention over time. From the visualization we can analyze the model's choice of attributes in the generated caption. We observe that the color, count, and size modules are more active at the beginning of a sentence and the initial estimation appears more dominant in the later half. More investigation will be needed to draw a conclusive explanation, but we hypothesize that 
it may be due to the fact that verbs and objects come first in the English language structure.  

Nonetheless, by explicitly proposing grounded attributes to the language model our model is able to include the proposed attributes in the target sentence more often and it is more likely to give ``risky'' but detailed descriptions of the content in an image.
\vspace{0pt}
\section{Conclusion}
In this work, we propose an image captioning model that utilizes neural module networks to propose specialized and grounded attributes. Experimental results show that our model achieves both the fluency of sequential models and the specificity of compositional models. Specifically, our approach excels at including fine-grained details such as counting that are generally avoided or overlooked. The framework is easily expandable to include additional functional modules of more sophisticated designs. Improved interpretability via visualized attention is another bonus because the model enables a quantitative analysis of both visual and semantic information.

\section*{Acknowledgements}
This work was conducted in part through
collaborative participation in the Robotics Consortium sponsored by
the U.S Army Research Laboratory under the Collaborative Technology
Alliance Program, Cooperative Agreement W911NF-10-2-0016. The views
and conclusions contained in this document are those of the authors
and should not be interpreted as representing the official policies,
either expressed or implied, of the Army Research Laboratory of the
U.S. Government. The U.S. Government is authorized to reproduce and
distribute reprints for Government purposes notwithstanding any
copyright notation herein.

\bibliographystyle{named}
\bibliography{ijcai19}

\begin{thebibliography}{}

\bibitem[\protect\citeauthoryear{Anderson \bgroup \em et al.\egroup
  }{2016}]{anderson2016spice}
Peter Anderson, Basura Fernando, Mark Johnson, and Stephen Gould.
\newblock Spice: Semantic propositional image caption evaluation.
\newblock In {\em European Conference on Computer Vision}, pages 382--398.
  Springer, 2016.

\bibitem[\protect\citeauthoryear{Anderson \bgroup \em et al.\egroup
  }{2017}]{anderson2017bottom}
Peter Anderson, Xiaodong He, Chris Buehler, Damien Teney, Mark Johnson, Stephen
  Gould, and Lei Zhang.
\newblock Bottom-up and top-down attention for image captioning and vqa.
\newblock {\em arXiv preprint arXiv:1707.07998}, 2017.

\bibitem[\protect\citeauthoryear{Andreas \bgroup \em et al.\egroup
  }{2016}]{andreas2016learning}
Jacob Andreas, Marcus Rohrbach, Trevor Darrell, and Dan Klein.
\newblock Learning to compose neural networks for question answering.
\newblock {\em arXiv preprint arXiv:1601.01705}, 2016.

\bibitem[\protect\citeauthoryear{Dai \bgroup \em et al.\egroup
  }{2017}]{dai2017detecting}
Bo~Dai, Yuqi Zhang, and Dahua Lin.
\newblock Detecting visual relationships with deep relational networks.
\newblock In {\em 2017 IEEE Conference on Computer Vision and Pattern
  Recognition (CVPR)}, pages 3298--3308. IEEE, 2017.

\bibitem[\protect\citeauthoryear{Dai \bgroup \em et al.\egroup
  }{2018}]{dai2018neural}
Bo~Dai, Sanja Fidler, and Dahua Lin.
\newblock A neural compositional paradigm for image captioning.
\newblock In {\em Advances in Neural Information Processing Systems}, pages
  656--666, 2018.

\bibitem[\protect\citeauthoryear{Denkowski and
  Lavie}{2014}]{denkowski2014meteor}
Michael Denkowski and Alon Lavie.
\newblock Meteor universal: Language specific translation evaluation for any
  target language.
\newblock In {\em Proceedings of the ninth workshop on statistical machine
  translation}, pages 376--380, 2014.

\bibitem[\protect\citeauthoryear{Fang \bgroup \em et al.\egroup
  }{2015}]{fang2015captions}
Hao Fang, Saurabh Gupta, Forrest Iandola, Rupesh~K Srivastava, Li~Deng, Piotr
  Doll{\'a}r, Jianfeng Gao, Xiaodong He, Margaret Mitchell, John~C Platt,
  et~al.
\newblock From captions to visual concepts and back.
\newblock In {\em Proceedings of the IEEE conference on computer vision and
  pattern recognition}, pages 1473--1482, 2015.

\bibitem[\protect\citeauthoryear{He \bgroup \em et al.\egroup
  }{2016}]{he2016deep}
Kaiming He, Xiangyu Zhang, Shaoqing Ren, and Jian Sun.
\newblock Deep residual learning for image recognition.
\newblock In {\em Proceedings of the IEEE conference on computer vision and
  pattern recognition}, pages 770--778, 2016.

\bibitem[\protect\citeauthoryear{Ilse \bgroup \em et al.\egroup
  }{2018}]{ilse2018attention}
Maximilian Ilse, Jakub~M Tomczak, and Max Welling.
\newblock Attention-based deep multiple instance learning.
\newblock {\em arXiv preprint arXiv:1802.04712}, 2018.

\bibitem[\protect\citeauthoryear{Karpathy and Fei-Fei}{2015}]{karpathy2015deep}
Andrej Karpathy and Li~Fei-Fei.
\newblock Deep visual-semantic alignments for generating image descriptions.
\newblock In {\em Proceedings of the IEEE conference on computer vision and
  pattern recognition}, pages 3128--3137, 2015.

\bibitem[\protect\citeauthoryear{Kingma and Ba}{2014}]{kingma2014adam}
Diederik~P Kingma and Jimmy Ba.
\newblock Adam: A method for stochastic optimization.
\newblock {\em arXiv preprint arXiv:1412.6980}, 2014.

\bibitem[\protect\citeauthoryear{Krishna \bgroup \em et al.\egroup
  }{2017}]{krishna2017visual}
Ranjay Krishna, Yuke Zhu, Oliver Groth, Justin Johnson, Kenji Hata, Joshua
  Kravitz, Stephanie Chen, Yannis Kalantidis, Li-Jia Li, David~A Shamma, et~al.
\newblock Visual genome: Connecting language and vision using crowdsourced
  dense image annotations.
\newblock {\em International Journal of Computer Vision}, 123(1):32--73, 2017.

\bibitem[\protect\citeauthoryear{Lin \bgroup \em et al.\egroup
  }{2014}]{lin2014microsoft}
Tsung-Yi Lin, Michael Maire, Serge Belongie, James Hays, Pietro Perona, Deva
  Ramanan, Piotr Doll{\'a}r, and C~Lawrence Zitnick.
\newblock {Microsoft COCO: Common objects in context}.
\newblock In {\em European conference on computer vision}, pages 740--755.
  Springer, 2014.

\bibitem[\protect\citeauthoryear{Lin}{2004}]{lin2004rouge}
Chin-Yew Lin.
\newblock Rouge: A package for automatic evaluation of summaries.
\newblock {\em Text Summarization Branches Out}, 2004.

\bibitem[\protect\citeauthoryear{Lu \bgroup \em et al.\egroup
  }{2017}]{lu2017knowing}
Jiasen Lu, Caiming Xiong, Devi Parikh, and Richard Socher.
\newblock Knowing when to look: Adaptive attention via a visual sentinel for
  image captioning.
\newblock In {\em Proceedings of the IEEE conference on computer vision and
  pattern recognition}, pages 375--383, 2017.

\bibitem[\protect\citeauthoryear{Papineni \bgroup \em et al.\egroup
  }{2002}]{papineni2002bleu}
Kishore Papineni, Salim Roukos, Todd Ward, and Wei-Jing Zhu.
\newblock Bleu: a method for automatic evaluation of machine translation.
\newblock In {\em Proceedings of the 40th annual meeting on association for
  computational linguistics}, pages 311--318. Association for Computational
  Linguistics, 2002.

\bibitem[\protect\citeauthoryear{Pennington \bgroup \em et al.\egroup
  }{2014}]{pennington2014glove}
Jeffrey Pennington, Richard Socher, and Christopher Manning.
\newblock Glove: Global vectors for word representation.
\newblock In {\em Proceedings of the 2014 conference on empirical methods in
  natural language processing (EMNLP)}, pages 1532--1543, 2014.

\bibitem[\protect\citeauthoryear{Ren \bgroup \em et al.\egroup
  }{2015}]{ren2015faster}
Shaoqing Ren, Kaiming He, Ross Girshick, and Jian Sun.
\newblock Faster r-cnn: Towards real-time object detection with region proposal
  networks.
\newblock In {\em Advances in neural information processing systems}, pages
  91--99, 2015.

\bibitem[\protect\citeauthoryear{Vedantam \bgroup \em et al.\egroup
  }{2015}]{vedantam2015cider}
Ramakrishna Vedantam, C~Lawrence~Zitnick, and Devi Parikh.
\newblock Cider: Consensus-based image description evaluation.
\newblock In {\em Proceedings of the IEEE conference on computer vision and
  pattern recognition}, pages 4566--4575, 2015.

\bibitem[\protect\citeauthoryear{Vinyals \bgroup \em et al.\egroup
  }{2015}]{vinyals2015show}
Oriol Vinyals, Alexander Toshev, Samy Bengio, and Dumitru Erhan.
\newblock Show and tell: A neural image caption generator.
\newblock In {\em Computer Vision and Pattern Recognition (CVPR), 2015 IEEE
  Conference on}, pages 3156--3164. IEEE, 2015.

\bibitem[\protect\citeauthoryear{Wang \bgroup \em et al.\egroup
  }{2017}]{wang2017skeleton}
Yufei Wang, Zhe Lin, Xiaohui Shen, Scott Cohen, and Garrison~W Cottrell.
\newblock Skeleton key: Image captioning by skeleton-attribute decomposition.
\newblock In {\em Computer Vision and Pattern Recognition (CVPR), 2017 IEEE
  Conference on}, pages 7378--7387. IEEE, 2017.

\bibitem[\protect\citeauthoryear{Yao \bgroup \em et al.\egroup
  }{2017}]{yao2017boosting}
Ting Yao, Yingwei Pan, Yehao Li, Zhaofan Qiu, and Tao Mei.
\newblock Boosting image captioning with attributes.
\newblock In {\em Proceedings of the IEEE International Conference on Computer
  Vision}, pages 4894--4902, 2017.

\bibitem[\protect\citeauthoryear{Yao \bgroup \em et al.\egroup
  }{2018}]{yao2018exploring}
Ting Yao, Yingwei Pan, Yehao Li, and Tao Mei.
\newblock Exploring visual relationship for image captioning.
\newblock In {\em Proceedings of the European Conference on Computer Vision
  (ECCV)}, pages 684--699, 2018.

\bibitem[\protect\citeauthoryear{You \bgroup \em et al.\egroup
  }{2016}]{you2016image}
Quanzeng You, Hailin Jin, Zhaowen Wang, Chen Fang, and Jiebo Luo.
\newblock Image captioning with semantic attention.
\newblock In {\em Proceedings of the IEEE Conference on Computer Vision and
  Pattern Recognition}, pages 4651--4659, 2016.

\bibitem[\protect\citeauthoryear{Zhang \bgroup \em et al.\egroup
  }{2006}]{viola2016-mil-noisy}
Cha Zhang, John~C. Platt, and Paul~A. Viola.
\newblock Multiple instance boosting for object detection.
\newblock In Y.~Weiss, B.~Sch\"{o}lkopf, and J.~C. Platt, editors, {\em
  Advances in Neural Information Processing Systems}, pages 1417--1424. MIT
  Press, 2006.

\end{thebibliography}
\end{document}